\documentclass[conference]{IEEEtran}
\usepackage{times}
\usepackage{graphicx}
\usepackage{subcaption}
\usepackage{graphicx}
\usepackage{xcolor}
\usepackage{amsmath}
\usepackage{amsfonts}
\usepackage{multirow}
\usepackage[]{units}

\usepackage[numbers]{natbib}
\usepackage{multicol}
\usepackage[bookmarks=true]{hyperref}
\usepackage{booktabs}

\newif\ifsubmit
\submittrue
\ifsubmit
\newcommand{\stnote}[1]{}
\newcommand{\ngnote}[1]{}
\newcommand{\lswnote}[1]{}
\newcommand{\sknote}[1]{}
\newcommand{\danote}[1]{}
\else
\newcommand{\stnote}[1]{\textcolor{blue}{\textbf{ST: #1}}}
\newcommand{\ngnote}[1]{\textcolor{green}{\textbf{NG: #1}}}
\newcommand{\lswnote}[1]{\textcolor{violet}{\textbf{LSW: #1}}}
\newcommand{\sknote}[1]{\textcolor{red}{\textbf{SK: #1}}}
\newcommand{\danote}[1]{\textcolor{olive}{\textbf{DA: #1}}}
\fi

\begin{document}
% \title{Grounding Language to Reward Functions With Abstract Markov Decision Processes}
%\title{Efficient Language Grounding and Execution with Hierarchical Planners}
\title{Efficiently and Accurately Interpreting Language at Multiple Levels of Abstraction}
\title{Accurately and Efficiently Interpreting Language Commands at Multiple Levels of Abstraction}
\title{Accurately and Efficiently Interpreting Human-Robot Commands at Multiple Levels of Abstraction}
\title{Efficiently Interpreting Human-Robot Instructions by Leveraging Hierarchical Structure in Tasks}
\title{Hierarchical Planning for Accurate Interpretation and Robust Execution of Human-Robot Instructions }
\title{Accurately and Efficiently Interpreting Human-Robot Instructions of Varying Granularities}
% \title{Accurately and Efficiently Interpreting Human-Robot Instructions at Multiple Levels of Abstraction}
% \author{Paper ID: 192}
%\author{Dilip Arumugam*
%\and Siddharth Karamcheti*
%\and Nakul Gopalan
%\and Lawson L.S. Wong
%\and Stefanie Tellex
%\thanks{* The first two authors contributed equally.}
%}
\author{\authorblockN{Dilip Arumugam\authorrefmark{1}, Siddharth Karamcheti\authorrefmark{1},
Nakul Gopalan, Lawson L.S. Wong, and Stefanie Tellex}
\authorblockA{Computer Science Department, Brown University, Providence, RI 02912\\
%dilip\_arumugam@brown.edu, siddharth\_karamcheti@brown.edu, ngopalan@cs.brown.edu, lsw@brown.edu, stefie10@cs.brown.edu\\
\{dilip\_arumugam@, siddharth\_karamcheti@, ngopalan@cs., lsw@, stefie10@cs.\}brown.edu\\
\authorrefmark{1} The first two authors contributed equally}
}

\maketitle
%\blfootnote{$^*$ Computer Science Department, Brown University \\
%$\dagger$ The first two authors contributed equally.
%}
%\renewcommand{\thefootnote}{\arabic{footnote}}
\begin{abstract}
% Humans can ground natural language commands to tasks at very high, abstract levels of specificity or very fine-grained, concrete levels. 
Humans can ground natural language commands to tasks at both abstract and fine-grained levels of specificity. For instance, a human forklift operator can be instructed to perform a high-level action, like ``grab a pallet'' or a low-level action like ``tilt back a little bit.'' While robots are also capable of grounding language commands to tasks, previous methods implicitly assume that all commands and tasks reside at a single, fixed level of abstraction. Additionally, methods that do not use multiple levels of abstraction encounter inefficient planning and execution times as they solve tasks at a single level of abstraction with large, intractable state-action spaces  closely resembling real world complexity. In this work, by grounding commands to all the tasks or subtasks available in a hierarchical planning framework, we arrive at a model capable of interpreting language at multiple levels of specificity ranging from coarse to more granular. We show that the accuracy of the grounding procedure is improved when simultaneously inferring the degree of abstraction in language used to communicate the task. Leveraging hierarchy also improves efficiency: our proposed approach enables a robot to respond to a command within one second on $90\%$ of our tasks, while baselines take over twenty seconds on half the tasks. Finally, we demonstrate that a real, physical robot can ground commands at multiple levels of abstraction allowing it to efficiently plan different subtasks within the same planning hierarchy.
\end{abstract}

\section{Introduction}
%\blfootnote{\\$^*$ Computer Science Department, Brown University \\
%$\dagger$ The first two authors contributed equally.
%}
% In order to facilitate seamless human-robot interaction, it is important for there to be minimal difference between human-human dialogue and human-robot dialogue. \stnote{Cut previous sentence; too abstract}
In everyday speech, humans use language at multiple levels of abstraction. For example, a brief transcript from an expert human forklift operator instructing a human trainee has very abstract commands such as ``Grab a pallet,'' mid-level commands such as ``Make sure your forks are centered,'' and very fine-grained commands such as ``Tilt back a little bit'' all
within thirty seconds of dialog.  Humans use these varied granularities to  
specify and reason about a large variety of tasks with a wide range of difficulties. Furthermore, these abstractions in language map to subgoals that are useful when interpreting and executing a task. In the case of the forklift trainee above, the sub-goals of moving to the pallet, placing the forks under the object, then lifting it up are all implicitly encoded in the command ``Grab a pallet.''  
%Yet when necessary, one can immediately drop to a finer level of abstraction by saying ``Turn slightly to the left.'' 
By decomposing generic, abstract commands into modular sub-goals, humans exert more organization, efficiency, and control in their planning and execution of tasks.  A robotic system that can identify and leverage the degree of specificity used to communicate instructions would be more accurate in its task grounding and more robust towards varied human communication.

% By inferring
% the latent degree of language specificity used to communicate the
% original task and incorporating an appropriate hierarchical planning
% framework, our work shows faster, improved accuracy in learning how to
% ground natural language commands to tasks defined at various levels of
% abstraction.  
 
%  \begin{figure}
%  \centering
%  \includegraphics[width=0.48\textwidth]{figures/example_turtlebot}
% \caption{The continuous object manipulation domain, where the Turtlebot agent needs to fetch a block into the goal room. \ngnote{need an example command here, may be a storyboard} \danote{Can we clarify what the purpose of this figure is supposed to be? Right now it seems misplaced }
% }
% \label{fig:tb1}
% \end{figure}
\begin{figure}
\centering
\includegraphics[width=.8\linewidth]{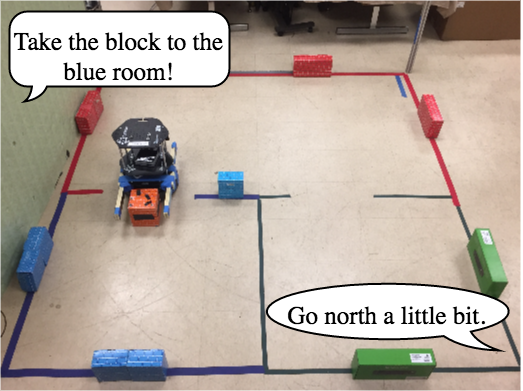}
\caption{Examples of high-level and fine-grained commands issued to the Turtlebot robot in a mobile-manipulation task.}
\vspace{-10pt}
\label{fig:tb1}
\end{figure}

Existing approaches map between natural language commands and a formal representation at some fixed level of abstraction~\citep{chen2011learning, matuszek2013learning, tellex11}.
% \stnote{Cut... Too wordy, and also don't want to be too judgemental about previous approaches (``cumbersome''): While there is prior work in the area of mapping between natural language commands and reward functions most approaches often map to some cumbersome task representation (\textit{e.g.} a featurized representation that directly translates to robot actions). }
While effective at directing robots to complete predefined tasks, mapping to fixed sequences of robot actions is unreliable in changing or stochastic environments. Accordingly, \citet{macglashan2015grounding} decouple the problem and use a statistical language model to map between language and robot goals, expressed as reward functions in a Markov Decision Process (MDP).  Then, an arbitrary planner solves the MDP, resolving any environment-specific challenges with execution.
%incident to the execution of the grounded task. 
As a result, the learned language model can  transfer to other robots with different action sets so long as there is consistency in the task representation (\textit{i.e.}, reward functions). However, MDPs for complex, real-world environments face an inherent tradeoff between including low-level task representations and increasing the time needed to plan in the presence of both low- and high-level reward functions~\citep{gopalan17}. 

%A limitation of this approach is that it can only account for commands at a single level of granularity. In less tractable, real world environments, choosing not to leverage the full information contained in language may not harm the overall task grounding process but severely limits one's expressibility, and dramatically increases the time needed to plan and execute the grounded task. \stnote{Need to tighten previous sentences up.}
%\stnote{In MDPs that model real-world problems, there is an inherent tradeoff between including fine-grained reward functions that can map to fine-grained natural language but increase planning time when they are included in an MDP that also has very abstract high-level reward functions~\citep{Gopalan2016PlanningWA}.}

To address these problems, we present an approach for mapping natural language commands of varying complexities to reward functions at different levels of abstraction within a hierarchical planning framework.  This approach enables the system to quickly and accurately interpret both abstract and fine-grained commands.  Our system uses a deep neural network language model that maps natural language commands to the appropriate level of the planning hierarchy. By coupling abstraction-level inference with the overall grounding problem, we exploit the subsequent hierarchical planner to efficiently execute the grounded tasks.  To our knowledge, we are the first to contribute a system for grounding language at multiple levels of abstraction, as well as the first to contribute a deep learning system for improved robotic language understanding.

Our evaluation shows that deep neural network language models can infer reward functions  
% \ngnote{verify this claim, where did we do timing analysis for ibm vs single-rnn?} \sknote{IBM is slower than Single-RNN - IBM requires marginalizing over all alignments - significantly longer than a single forward pass.} 
more accurately than statistical language model baselines. We present results comparing a traditional statistical language model to three different neural architectures that are commonly used in natural language processing. Furthermore, we show that a hierarchical approach allows the planner to map to a larger, richer space of reward functions more quickly and more accurately than non-hierarchical baselines.  This speedup allows the robot to respond faster and more accurately to a user's request, with a much larger set of potential commands than previous approaches. We also demonstrate on a Turtlebot the rapid and accurate response of our system to natural language commands at varying levels of abstraction.

\section{Related work}
% Accordingly, emulating the same standard of communication between humans and robots has been a problem of great interest.
%\stnote{Cut: Consequently, natural language is a readily available tool for humans to communicate tasks and motivations to other types of partners they work with such as robots. }
%
% two sentence agenda
%\stnote{Cut (all this should be said in the intro already): A robot that needs to work with the general public needs the ability to understand natural language commands given by its human partners. 
%Further, the robot also needs to understand the level of abstraction with which a task is specified and then solve it. For example, a robot that can complete the task of ``walk to the next room'' should also be able to comprehend and solve the task of ``move forward two steps,'' much like a human who learns these tasks. }

% previous methods
Humans use natural language to communicate ideas, motivations, task descriptions, etc. with other humans.
Some of the earliest works in this area mapped tasks to another planning language, which then grounded to the actions performed by the robots \citep{Dzifcak2009WhatTD,chen2011learning}. 
More recent methods ground natural language commands to tasks using features that describe correspondences between natural language phrases present in the task description to the physical objects  \citep{howard2014natural, matuszek2013learning, tellex11,Brooks2012MakeIS,Raman2011AnalyzingUS}, or abstract spatial concepts \cite{Paul2016abstract}, present in the world and the actions available in the world. This featurized representation can then describe the sequence of actions needed to complete the task.
%using feature representations that describe the sequence of actions that need to be performed to complete the tasks [,  Cynthia's previous work].  
% more recent methods like Rohan Paul that grounds abstract ideas like rows and columns, an improvement on this idea
%In a similar vein, \citet{Paul2016abstract} ground to abstract spatial concepts like \emph{rows}, \emph{columns}, and \emph{middle} before learning correspondences between them to solve tasks. 
All these approaches ground commands to action sequences, leading to brittle behavior if the environment is stochastic.

%The reward functions were learned using trajectory demonstrations via IRL \citep{ng2000algorithms}.
% James's idea

\citet{macglashan2015grounding} proposed grounding natural language commands to reward functions associated with certain tasks, allowing robot agents to plan in stochastic environments.
%\sknote{Is this necessary?: The robot can solve for individual plans once natural language commands ground to applicable reward functions.} 
They treat the goal reward function as a sequence of propositional functions, much like a machine language, to which a natural language task can be translated, using an IBM Model 2 \cite{Brown1990ASA,Brown1993TheMO} (IBM2) language model. While their propositional functions only lie at one level of abstraction, we want the robot to understand commands at different levels of specificity while still maintaining efficient planning and execution in the face of multiple levels of abstraction. 

Crucially, \citet{macglashan2015grounding} actually perform inference over reward function templates, or lifted reward functions, along with environmental constraints. A lifted reward function merely specifies a task while leaving the environment-specific variables of the task undefined. The environmental binding constraints then specify the properties that an object in the environment must satisfy in order to be bound to a lifted reward function variable. By doing this, the output space of the language model is never tied to any particular instantiation of the environment, but can instead align to objects and attributes that lie within some distribution over environments. Given a lifted reward function and environment constraints (henceforth jointly referred to as only a lifted reward function), a subsequent model can later infer the environment-specific variables without needing to relearn the language understanding components for each environment. In order to leverage this flexibility, all of our proposed language models  produce lifted reward functions which are then completed by a grounding module before being passed to the planner (see Sec.~\ref{sec:lm}).
%, that is, on the level of grid cells, to rooms, to buildings, to cities while still maintaining timely planning and execution.   \stnote{The previous work cannot do this because the planner cannot efficiently plan with these different levels of abstraction.}
%This means that if the robot were trained to move forward a step and go to different rooms, it would solve the tasks at the same flat level of abstraction. 

Planning in domains with large state-action spaces is computationally expensive as planners like value iteration and bounded real-time dynamic programming (RTDP) need to explore the domain at the lowest, ``flat'' level of abstraction \citep{Bellman:1957,McMahan2005BoundedRD}.
Naively this might result in an exhaustive search of the space before the goal state is found. 
A better approach is to decompose the planning problem into smaller, more easily solved subtasks. 
%where a subtask is a sequence of actions used to complete a subgoal.
The agent can then achieve the goal by choosing a sequence of these subtasks.
A common method to describe subtasks is by using temporal abstraction in the form of macro-actions \citep{mcgovern1997roles} or options \citep{Sutton1999BetweenMA}. 
These methods achieve subgoals using either a fixed sequence of actions \citep{mcgovern1997roles} or a subgoal based policy \citep{Sutton1999BetweenMA}. 
Planning with macro-actions or options requires computing the policies for each option or macro-action, which is done by exploring and backing up rewards from lowest level actions. 
This ``bottom-up'' planning is slow, as the reward for each action taken needs to be backed up through the hierarchy of options, which is time consuming.
Other methods for abstraction, like MAXQ \citep{Dietterich2000}, R-MAXQ \citep{jong2008hierarchicalRMAXQ} and Abstract Markov Decision Processes (AMDPs) \citep{gopalan17} involve providing a hierarchy of subtasks.
In these methods, a subtask is associated with a subgoal and a state abstraction relevant to achieving the subgoal \citep{Dietterich2000,gopalan17,jong2008hierarchicalRMAXQ}.
Both MAXQ \citep{Dietterich2000} and R-MAXQ \citep{jong2008hierarchicalRMAXQ} are bottom-up planners, they back up each individual action's reward across the hierarchy. 

We use AMDPs~\citep{gopalan17} in this paper because they plan in a ``top-down'' fashion. AMDPs offer model-based hierarchical representations in the form of reward functions and transition functions to every subtask. An AMDP hierarchy itself is an acyclic graph in which each node is a primitive action or an AMDP that solves a subtask defined by its parent; the states of each subtask AMDP are abstract representations of the environment state. AMDPs have been shown to achieve faster planning performance than other hierarchical methods \cite{gopalan17}.

%\ngnote{this sentence needs clarification - I want to say that it is not really possible in MAXQ to ground a single subtask without querying the root and solving the entire problem description.} 

%\sknote{I didn't know how much to add, so I covered the basics regarding word embeddings/RNNs here, but not in mathematical detail - that is still in the later sections}
We use a deep neural network language model to perform language grounding. Deep neural networks have had great success in many natural language processing (NLP) tasks, such as traditional language modeling \citep{Bengio2000ANP,Mikolov2010RecurrentNN,Mikolov2011ExtensionsOR}, machine translation~\citep{Cho2014LearningPR,Chung2014EmpiricalEO}, and text categorization~\citep{Iyyer2015DeepUC}.
%One of the contributing factors to the success of such methods is
One reason for their success is the ability to learn meaningful input representations~\citep{Bengio2000ANP,Mikolov2013EfficientEO}.
%LSW% For example, \citet{Bengio2000ANP} learn distributed representations of words in tandem with the rest of their language model. Similarly, \citet{Mikolov2013EfficientEO} propose a system solely dedicated to learning these representations.
These ``embeddings'' are dense vectors that not only uniquely represent individual words (as opposed to otherwise sparse approaches for word representation), but also capture semantically significant features of the language.
%Another contributing factor to the success of such methods
Another reason is the use of recurrent neural networks (RNNs), a type of neural network cell that maps variable length inputs (i.e. commands) to a fixed-size vector representation, which have been widely used in NLP~\citep{Cho2014LearningPR,Chung2014EmpiricalEO,Yamada2016DynamicalLO}.
% To the best of our knowledge, our approach is the first to
Our approach uses both word embeddings and a state-of-the-art RNN model to map between natural language and MDP reward functions.

% In addition, Recurrent Neural Networks (RNNs) have been proven to be extraordinarily useful for a wide-variety of sequence-based natural language tasks, like translation and language modeling. Specifically, we utilize the Gated Recurrent Unit, a type of RNN that has been proven effective for problems in machine translation (\cite{Cho2014LearningPR,Chung2014EmpiricalEO}). Whereas a traditional feed-forward neural network is characterized by a series of transformations to a fixed-sized input, RNNs are characterized by their ability to transform variable-length sequences. Specifically, RNNs are defined by a hidden state vector, that is updated as new inputs are received. By persisting the hidden state through entire sequences, RNNs are able to learn robust representations of entire sequences, which can then be fed to subsequent network layers. 

\section{Technical Approach}

To interpret a variety of natural language commands, there must be a representation for all possible tasks and subtasks.  We specify an \textit{Object-oriented Markov Decision Process} (OO-MDP) to model the robot's environment and actions~\citep{Diuk2008AnOR}.
An MDP is a five-tuple of $\langle \mathcal{S}, \mathcal{A}, \mathcal{T}, \mathcal{R}, \gamma \rangle$ where $\mathcal{S}$ represents the set of states that define an environment, $\mathcal{A}$ denotes the set of actions an agent can execute to transition between states, $\mathcal{T}$ defines the transition probability distribution over all possible next states given a current state and executed action, $\mathcal{R}$ defines the numerical reward earned for a particular transition, and $\gamma$ represents the discount factor or effective time horizon under consideration. Planning in an MDP produces a mapping between states and actions, or policy, that maximizes the total expected discounted reward.   In our framework, as in \citet{macglashan2015grounding}, we will map between words in language and specific reward functions. 

An OO-MDP builds upon an MDP by adding sets of object classes and propositional functions; each object class is defined by a set of attributes and each propositional function is parameterized by instances of object classes.
%By virtue of using an OO-MDP, every pertinent object in the state of a domain is given its own unique identifier, which then appears alongside the high level symbolic information encoded by the propositional functions.
For example, an OO-MDP for the mobile robot manipulation domain seen in Fig.~\ref{fig:tb1} might denote the robot's successful placement of the orange block into the blue room via the propositional function {\sf blockInRoom block0 room1}, where {\sf block0} and {\sf room1} are instances of the block and room object classes respectively and the {\sf blockInRoom} propositional function checks if the location attribute of {\sf block0} is contained in {\sf room1}. Using these propositional functions as reward functions that encode termination conditions for each task, we arrive at a sufficient, semantic representation for grounding language. For our evaluation, we use the Cleanup World~\citep{Junghanns1997SokobanAC,macglashan2015grounding} OO-MDP,  which models a mobile manipulator robot; this domain is defined in Sec.~\ref{sec:cw}.

However, this approach does not generalize well to different environment configurations. At training time, any natural language command that moves objects or agents to 
a specific room is conditioned to map room attributes to specific room instances (i.e. in the case of Fig.~\ref{fig:tb1}, the blue room is always {\sf room1}). With this in mind, consider what happens if we switched the blue and green rooms at test time, so that the green room is now {\sf room1}. In this case, any language command that moves an object or agent to the blue room would fail, as the room instances have been switched around.

To this end, we ``lift'' the propositional functions from before, to better generalize to unseen environments. Given a command like ``Take the block to blue room,'' the corresponding lifted propositional function takes the form {\sf blockInRoom block0 roomIsBlue}, denoting that the block should end up in the room that is blue. We then assume an environment-specific grounding module (see Sec.~\ref{sec:gm}) that consumes these lifted reward functions and performs the actual low-level binding to specific room instances, which can then be passed to a planner.

In order to effectively ground commands across multiple levels of complexity, we assume a predefined hierarchy over the state-action space of the given grounding environment.  Furthermore, each level of this hierarchy requires its own set of reward functions for all relevant tasks and sub-tasks.  In our work, fast planning and the ability to ground and solve individual subtasks without needing to solve the entire planning problem make AMDPs a reliable choice for the hierarchical planner~\citep{gopalan17}. Finally, we assume that all commands are generated from a single, fixed level of abstraction.

Given a natural language command $c$, we find the corresponding level of the abstraction hierarchy $l$, and the lifted reward function $m$ that maximizes the joint probability of $l$, $m$ given $c$. Concretely, we seek the level of the state-action hierarchy $\hat l$ and the lifted reward function $\hat m$ such that:
\begin{align} 
	\hat l, \hat m &= \arg \max_{l, m} Pr(l, m \mid c) \label{eq:CoreObj}
\end{align}
For example, as illustrated in Fig.~\ref{fig:tb1}, a high-level natural language command like ``Take the block to the blue room'' would map to the highest abstraction level, while a low-level command like ``Go north a little bit'' would map to the finest-grained level.  We estimate this joint probability by learning a language model (described in Sec.~\ref{sec:lm}) and training on a parallel corpus that pairs natural language commands with a corresponding reward function at a particular level of the abstraction hierarchy.

%\subsection{Learning Models}

%The learning process for each model finds the model parameters that maximize likelihood of a training corpus. The process begins with a parallel corpus, commonly used in translation literature, with natural language commands mapped to the corresponding level of the state-action hierarchy, as well as to the corresponding reward function. Note that we assume each individual natural language command is generated at exactly one level of abstraction. \sknote{Assumption!} 

%This parallel corpus is collected via Amazon Mechanical Turk, and the specific details of the collection process is discussed in section \ref{sec:er}.

Given this parallel corpus, we train each model by directly maximizing the joint probability from Eqn.~\ref{eq:CoreObj}. Specifically, we learn parameters $\hat \theta$ that maximize the corpus likelihood:
\begin{align}
	\hat \theta &= \arg \max_{\theta} \prod_{(c, l, m) \in \mathbb{C}} Pr(l, m \mid c, \theta) \label{eq:CorpusObj}
\end{align}
At inference time, given a language command $c$, we find the best $l$, $m$ that maximize the probability $Pr(l, m \mid c, \hat \theta)$. The lifted reward function $m$ is then completed by the grounding module (see Sec.~\ref{sec:gm}) and passed to a hierarchical planner, which plans the corresponding task at abstraction level $l$.

\begin{figure*}
\centering
\subcaptionbox{Multi-NN Model\label{fig:multinn}}{
\includegraphics[height=.2\textheight]{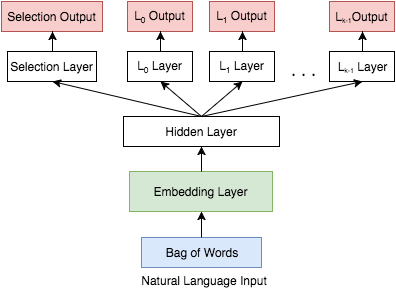}}
\hfill
\subcaptionbox{Multi-RNN Model\label{fig:multirnn}}{
\includegraphics[height=.2\textheight]{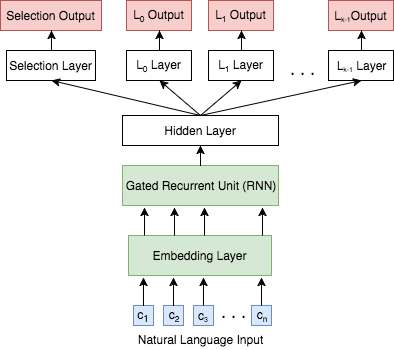}}
\hfill
\subcaptionbox{Single-RNN Model\label{fig:singlernn}}{
\includegraphics[height=.2\textheight]{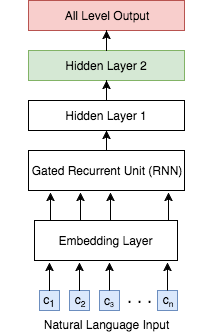}}
\caption{Model architectures for all three sets of deep neural network models. In blue are the network inputs, and in red are the network outputs. Going left to right, the green denotes significant structural differences between models.}
\label{fig:architectures}
\vspace{-10pt}
\end{figure*}

\section{Language Models}
\label{sec:lm}
%\sknote{Removed: Making use of both statistical and deep neural network language models, we examined the success of language groundings at multiple levels of specificity. A breakdown of the different models is as follows:}

We compare four language models: an IBM Model 2 translation model (similar to \citet{macglashan2015grounding}), a deep neural network bag-of-words language model, and two recurrent neural network (RNN) language models, with varying architectures. For detailed descriptions and implementations of all the presented models, as well as the datasets used throughout this paper, please refer to the supplemental repository: \\
\indent \href{https://github.com/h2r/GLAMDP}{https://github.com/h2r/GLAMDP}.

\subsection{IBM Model 2}

%\stnote{You can probably cut this whole section and just say you followed Macglashan et al, and also the details of the training.... if there is need for more space.}

As a baseline, task grounding is formulated as a machine translation problem, with natural language as the source language and semantic task representations (lifted reward functions) as the target language. We use the well-known IBM Model 2 (IBM2) machine translation model \cite{Brown1990ASA,Brown1993TheMO} as a statistical language model for scoring reward functions given input commands. 
% Following from equation \ref{eq:CoreObj} , we formalize the problem as follows: Given a natural language command $c$, we wish to find a level of abstraction $\hat l$ and a reward function $\hat m$ that maximizes the following probability:
IBM2 is a generative model that solves the following objective (equivalent to Eqn.~\ref{eq:CoreObj} by Bayes' rule):
\begin{align}
%	\hat l, \hat m &= \arg \max_{l, m} Pr(l, m \mid c) \\
       \hat l, \hat m  &= \arg \max_{l, m} Pr(l, m) \cdot Pr (c \mid l, m) \label{eq:Noisy-Channel}
\end{align}

This task grounding formulation follows directly from \citet{macglashan2015grounding} and we continue in an identical fashion training the IBM2 using the standard EM algorithm.

\subsection{Neural Network Language Models}

% \stnote{Cut:  (already said it in releated work)Deep neural networks have demonstrated robust performance on several problems in vision, speech, and natural language processing \citep{Krizhevsky2012ImageNetCW,Simonyan2014VeryDC,Hinton2012DeepNN,Graves2013SpeechRW,Cho2014LearningPR,Bengio2000ANP}. Inspired by this success,} 

We develop three classes of neural network architectures (see Fig.~\ref{fig:architectures}): a feed-forward network that takes a natural language command encoded as a bag-of-words and has separate parameters for each level of abstraction (\textbf{Multi-NN}), a recurrent network that takes into account the order of words in the sequence, also with separate parameters (\textbf{Multi-RNN}), and a recurrent network that takes into account the order of words in the sequence and has a shared parameter space across levels of abstraction (\textbf{Single-RNN}).
\bigskip

% To train these models, we use triples $\{c, l, m\}$, where $c$ denotes the input natural language command, $l$ denotes the correct AMDP level of abstraction for the given command, and $m$ denotes the correct reward function for the given command. We directly maximize the joint probability of the level $l$ and reward function $m$, given the natural language command $c$. More precisely, we wish to find a set of network parameters $\hat \theta$, such that:
% \begin{align} \label{align:NN}
% 	\hat \theta &= \arg \max_{\theta} \sum_{(c, l, m)} 
%                   \log Pr(l, m \mid c)   
% \end{align}
% % &= \arg \max_{\theta} \sum_{(c, l, m) \in N} \log Pr(l \mid c) + \log Pr(m \mid l, c)

\subsubsection{\textbf{Multi-NN: Multiple Output Feed-Forward Network}}
We propose a feed-forward neural network~\citep{Bengio2000ANP,Iyyer2015DeepUC,Mikolov2013EfficientEO} that takes in a natural language command $c$ as a bag-of-words vector $\vec{c}$, and outputs both the probability of each of the different levels of abstraction, as well as the probability of each reward function. We decompose the conditional probability from 
Eqn.~\ref{eq:CoreObj} as $Pr(l, m \mid c) = Pr(l \mid c) \cdot Pr(m \mid l, c)$.
Applying this to the corpus likelihood (Eqn.~\ref{eq:CorpusObj}) and taking logarithms,
the Multi-NN objective is to find parameters $\hat \theta$: %such that:
\begin{align} \label{align:MultiNN}
	\hat \theta &= \arg \max_{\theta} \sum_{(\vec{c}, l, m)} \log Pr(l \mid \vec{c}, \theta) + \log Pr(m \mid l, \vec{c}, \theta)
\end{align}
%This follows by taking a logarithm of the corpus objective outlined in Eqn.~\ref{eq:CorpusObj}.

To learn this set of parameters, we use the architecture shown in Fig.~\ref{fig:multinn}. Namely, we employ a multi-output deep neural network with an initial embedding layer, a hidden layer that is shared between each of the different outputs, and then output-specific hidden and read-out layers, respectively.

The level selection output is a $k$-element discrete distribution, where $k$ is the number of levels of abstraction in the given planning hierarchy. Similarly, the reward function output at each level $L_i$ is an $r_i$-element distribution, where $r_i$ is the number of reward functions at the given level of the hierarchy.
%(each of the reward functions at a specific level of the AMDP is given a unique integer identifier, held constant throughout training and testing).

To train the model, we minimize the sum of the cross-entropy loss on each term in Eqn.~\ref{align:MultiNN}. We train the network via backpropagation, using the Adam Optimizer \citep{Kingma2014AdamAM}, with a mini-batch size of 16, and a learning rate of 0.001. Furthermore, to better regularize the model and encourage robustness, we use Dropout \citep{Srivastava2014DropoutAS} after the initial embedding layer, as well as after the output-specific hidden layers with probability $p=0.5$.

\subsubsection{\textbf{Multi-RNN: Multiple Output Recurrent Network}}

%\stnote{Cut: Recurrent neural networks hold state-of-the-art results in several natural language processing domains, and are the staple of sequence modeling tasks like language modeling and machine translation \citep{Cho2014LearningPR,Mikolov2010RecurrentNN,Mikolov2011ExtensionsOR,Sutskever2014SequenceTS}. }

Inspired by the success of recurrent neural networks (RNNs) in NLP tasks \citep{Cho2014LearningPR,Mikolov2010RecurrentNN,Mikolov2011ExtensionsOR,Sutskever2014SequenceTS}, we propose an RNN language model that takes in a command as a sequence of words and, like the Multi-NN bag-of-words model, outputs both the probability of each of the different levels of abstraction, as well as the probability of each reward function, at each level of abstraction. RNNs extend feed-forward networks to handle variable length inputs by employing a set of one or more hidden states, which are updated after reading in each input token. 
Instead of converting natural language command $c$ to a vector $\vec{c}$, we use an RNN to interpret it as a sequence of words $s = \langle c_1, c_2 \ldots c_n \rangle$.
% Similar to Multi-NN, we decompose the conditional probability from the objective in Eqn.~\ref{eq:CoreObj}. If we represent the natural language command $c$ as a sequence of words $s = \langle c_1, c_2 \ldots c_n \rangle$, the Multi-RNN objective is to find a set of parameters such that:
The Multi-RNN objective is then:
\begin{align} \label{align:MultiRNN}
	\hat \theta = \arg \max_{\theta} \sum_{(c, l, m)} &\log Pr(l \mid s, \theta) + \log Pr(m \mid l, s, \theta)
\end{align}

This modification is reflected in Fig.~\ref{fig:multirnn}, which is similar to the Multi-NN architecture, except in the lower layers where we use an RNN encoder that takes the sequence of raw input tokens
%(in lieu of a bag-of-words representation),
and maps them into a fixed-size state vector.
We use the gated recurrent unit (GRU) of \citet{Cho2014LearningPR}, a particular type of RNN cell
that have been shown to work well on natural language sequence modeling tasks \cite{Chung2014EmpiricalEO}.

%To do this, we use the architecture depicted in Fig.~\ref{fig:multirnn}; 
%\stnote{Note how I fixed the figure label - you can directly ref a subfigure and it will do the right thing.} 
%similar to the Multi-NN architecture, we instead use a Recurrent Neural Network encoder that takes the sequence of raw input tokens (in lieu of a bag-of-words representation), and maps them into a fixed-size state vector. 

%They do this by employing a set of one or more hidden states $h_t$ whose activation at a given time step $t$ (or in our case, word index) is dependent on the hidden state $h_{t - 1}$, as well as the input $x_t$, combined via a set of update rules. 

%We use the gated recurrent unit (GRU) of \citet{Cho2014LearningPR}, a particular type of RNN cell that is characterized by a hidden state incrementally updated with new inputs (\textit{i.e.} words in a command). GRUs have been shown to work well on natural language sequence modeling tasks \cite{Chung2014EmpiricalEO}.

Similar to the Multi-NN, we train the model by minimizing the sum of the cross-entropy loss of each of the two terms in Eqn.~\ref{align:MultiRNN}, with the same optimizer setup as the Multi-NN model. Dropout is used to regularize the network after the initial embedding layer and the output-specific hidden layers.

\begin{figure*}[t]
  \centering
  \hfill
  \begin{subfigure}{.25\linewidth}
  	\centering
    \includegraphics[width=.65\linewidth]{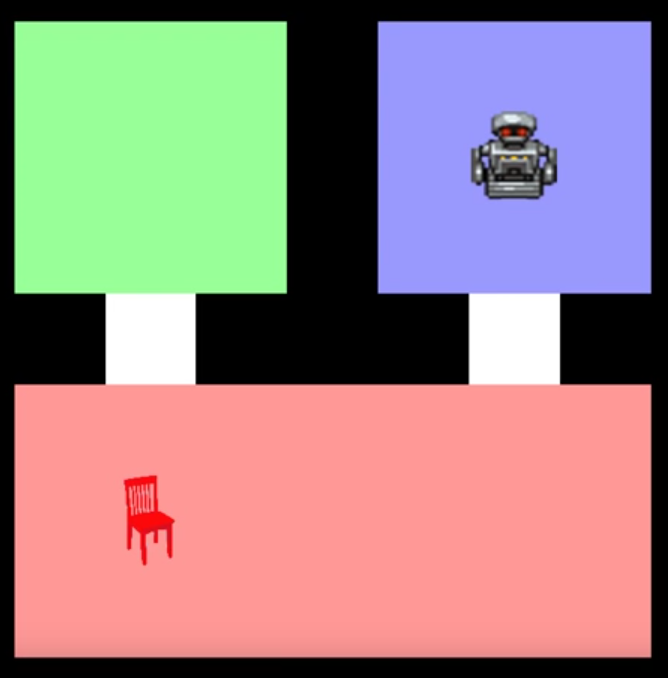}
 	\caption{A starting instance of the Cleanup World domain.}
 	\label{fig:cw1}
  \end{subfigure}
%   \begin{subfigure}{.25\linewidth}
%     \centering
%     %\begin{tabular}{l|c|c}
%      %       & Raw Data & Cleaned Data \\ \hline
%      % $L_0$ & 569      & 455          \\ %\hline
%      % $L_1$ & 572      & 341          \\ %\hline
%      % $L_2$ & 566      & 508          \\ %\hline
%     %\end{tabular}
%      \begin{tabular}{lr}
% \toprule
%             & Commands \\ \midrule
%       $L_0$ & 1309      \\ %\hline
%       $L_1$ & 872      \\ %\hline
%       $L_2$ & 866      \\ %\hline
% \bottomrule
%     \end{tabular}
%     \caption{Number of commands collected at each abstraction level. The $L_0$ corpus includes commands for moving one unit in each of the four cardinal directions which do not translate to or exist at higher levels \stnote{I would probably kill this table and give this information in the text when you describe the dataset collection.  Will save space.}}
%     \label{table:DataStats}
%   \end{subfigure}
  \hfill
  \begin{subfigure}{.7\linewidth}
    \centering
    {\small
    \begin{tabular}{lll}
\toprule
Level       &  Example Command   &  Reward Function \\
\midrule
      $L_0$ &
        \begin{tabular}[c]{@{}l@{}}
        	Turn and move one spot to the right. \\
        	Go three down, four over, two up.
        \end{tabular}
        &  
        \begin{tabular}[c]{@{}l@{}}
        	\sf{goWest} \\
        	\sf{agentInRoom agent0 roomIsGreen}
        \end{tabular} \\ 
         \hline
      $L_1$ &
        \begin{tabular}[c]{@{}l@{}}
        	Go to door, enter red room, \\ \hspace{8pt} push chair to green room door. \\
        	Go to the door then go into the red room.  \\
        \end{tabular} 
        &
        \begin{tabular}[c]{@{}l@{}}
        	\sf{blockInRegion block0 roomIsGreen} \\ \\
        	\sf{agentInRegion agent0 roomIsRed} \\
        \end{tabular} \\
        \hline
      $L_2$ &
        \begin{tabular}[c]{@{}l@{}}
        	Go to the green room. \\
          	Bring the chair to the blue room. 
        \end{tabular}
        & 
        \begin{tabular}[c]{@{}l@{}}
        	\sf{agentInRegion agent0 roomIsGreen} \\
        	\sf{blockInRegion block0 roomIsBlue}
        \end{tabular} \\
\bottomrule
    \end{tabular}
    }
    \caption{Example commands and corresponding reward functions. %\stnote{If you kill 3a,  you can make this table three tables (one for each level), all across the top.  That will save verticle space.   Also you can use a latex parbox inside table cells for more normal formatting.}
    }
    \label{table:Examples}
  \end{subfigure}
  \hfill
  \caption{Amazon Mechanical Turk (AMT) dataset domain and examples.}
  \label{figure:amt}
  \vspace{-10pt}
\end{figure*}

\subsubsection{\textbf{Single-RNN: Single Output Recurrent Network}}

Both Multi-NN and Multi-RNN decompose the conditional probability of both the level of abstraction $l$ and the lifted reward function $m$ given the natural language command $c$ as $Pr(l, m \mid c) = Pr(l \mid c) \cdot Pr(m \mid l, c)$, allowing for the explicit calculation of the probability of each level of abstraction given the natural language command. As a result, both Multi-NN and Multi-RNN create separate sets of parameters for each of the separate outputs, \textit{i.e.} separate parameters for each level of abstraction in the underlying hierarchical planner.

Alternatively, we can directly estimate the joint probability $Pr(l, m \mid c)$. To do so, we propose a different type of RNN model that takes in a natural language command as a sequence of words $s$ (as in Multi-RNN), and directly outputs the joint probability of each tuple $(l, m)$, where $l$ denotes the level of abstraction, and $m$ denotes the lifted reward function at the given level.
%Specifically, if we represent the natural language command $c$ as a sequence of words $s = \langle c_1, c_2 \ldots c_n \rangle$, then the Single-RNN objective is to find a set of parameters $\hat \theta$ such that:
The Single-RNN objective is to find $\hat \theta$ such that:
\begin{align} \label{align:SingleRNN}
	\hat \theta = \arg \max_{\theta} \sum_{(n, l, m)} \log Pr(l, m \mid s, \theta)
\end{align}

With this Single-RNN model, we are able to significantly improve model efficiency compared to the Multi-RNN model, as all levels of abstraction share a single set of parameters. Furthermore, removing the explicit calculation of the level selection probabilities allows for the possibility of positive information transfer between levels of abstraction, which is not necessarily possible with the previous models.

The Single-RNN architecture is shown in Fig.~\ref{fig:singlernn}. We use a single-output RNN, similar to the Multi-RNN architecture, with the key difference being that there is only a \emph{single} output, with each element of the final output vector corresponding to the probability of each tuple of levels of abstraction and reward functions $(l, m)$ given the natural language command $c$. 

% A detailed breakdown of the Single-RNN transformations are as follows:
% \begin{align*}
% 	\vec{e}_1, \vec{e}_2 \ldots \vec{e}_n &= \text{Lookup}(\mathbf{E}, c_1, c_2 \ldots c_n) \\
%     \vec{h} &= \text{GRU}(\vec{e}_1, \vec{e}_2, \ldots \vec{e}_n) \\
%     \vec{s} &= \text{ReLU}(\vec{h} \cdot \mathbf{W_s} + \mathbf{b}_s) \\
%     \vec{t} &= \text{ReLU}(\vec{s} \cdot \mathbf{W_t} + \mathbf{b}_t) \\
%     \vec{o} &= \text{Softmax}(\vec{t} \cdot \mathbf{W_t} + \mathbf{b}_t)
% \end{align*}
% Note that these transformations are exactly the same as the Multi-RNN, with the sole exception that there is only a single output, rather than multiple.

To train the model, we minimize the cross-entropy loss of the joint probability term in Eqn.~\ref{align:SingleRNN}. Training hyperparameters are identical to Multi-RNN, and dropout is applied to the initial embedding layer and the penultimate hidden layer.

\subsection{Grounding Module}
\label{sec:gm}

In all of our models, %we require an additional step of binding
the inferred lifted reward function template must be binded to environment-specific variables. The grounding module maps the lifted reward function to a grounded one that can be passed to an MDP planner. In our evaluation domain (see Fig.~\ref{fig:tb1}), it is sufficient for our grounding module to be a lookup table that maps specific environment constraints to object ID tokens. In domains with ambiguous constraints (\textit{e.g.} a ``chair'' argument where multiple chairs exist), a more complex grounding module could be substituted. For instance, \citet{artzi13} present a model for executing lambda-calculus expressions generated by a combinatory categorial grammar (CCG) semantic parser, which grounds ambiguous predicates and nested arguments. 

\section{Evaluation}
\label{sec:er}

% \begin{figure}[b]
%  \centering
%  \includegraphics[width=0.6\linewidth]{figures/cleanup_start}
%  \caption{A starting instance of the Cleanup World domain from which all AMT user study data was collected.}
%  \label{fig:cw1}
% \end{figure}

\begin{figure*}[t]
  \centering
  \begin{subfigure}{.48\linewidth}
  	\centering
    {\small
    \begin{tabular}{lrrr}
\toprule
                    &   Evaluated $L_0$         & Evaluated $L_1$        & Evaluated $L_2$   \\ \midrule
     Trained $L_0$  &   \boldmath$21.61\%$  & \boldmath$17.20\%$ & 21.87\%          \\ 
     Trained $L_1$  &   9.83\%            & 10.23\%           & 13.90\%           \\ 
     Trained $L_2$  &  14.94\%            & 12.84\%          & \boldmath$31.49\%$ \\
\bottomrule
    \end{tabular}
    }
    \caption{IBM2 Reward Grounding Baselines}
   	\label{results:baseIBM}
  \end{subfigure}
  \hfill
  \begin{subfigure}{.48\linewidth}
  	\centering
    {\small
    \begin{tabular}{lrrr}
\toprule
                    &   Evaluated $L_0$       & Evaluated $L_1$       & Evaluated $L_2$       \\ \midrule
     Trained $L_0$ &   \boldmath$77.67\%$ & 28.05\%          & 23.26\%          \\ 
     Trained $L_1$ &   32.79\%          & \boldmath$82.99\%$ & 74.65\%          \\ 
     Trained $L_2$ &   14.19\%          &   58.62\%        & \boldmath$87.91\%$\\ 
     %Trained $L_{ALL}$ model &   $82.9\%$          &   $88.4\%$        & $86.8\%$ \\ 
\bottomrule
    \end{tabular}
    }
    \caption{Single-RNN Reward Grounding Baselines}
   	\label{results:baseRNN}
  \end{subfigure}
  \hfill
  \caption{Task grounding accuracy (averaged over 5 trials) when training IBM2 and Single-RNN models on a single level
           of abstraction, then evaluating commands from alternate levels. This is similar to the 			                \citet{macglashan2015grounding} results, as we see that without accounting for abstractions in 	  			   language, there is a noticeable effect on grounding accuracy.}
  \vspace{-10pt}
\label{results:base}
\end{figure*}
\begin{figure}[t]
  \centering
  %\begin{tabular}{l|c|c||c|c}
  %& \multicolumn{2}{c||}{\textbf{Level Selection}} & \multicolumn{2}{c}{\textbf{Reward Grounding}} \\
  %            & Raw Dataset               & Cleaned Dataset          & Raw Dataset            & Cleaned Dataset           \\ \hline
  %IBM Model 2 & 77.44\%           & 85.81\%           & 23.95\%          & 25.42\%           \\
  %Multi-NN    & 97.67\%           & 97.21\%           & 37.54\%          & 38.01\%           \\
  %Multi-RNN   & 99.46\%           & \textbf{99.64\%}  & 80.85\%          & \textbf{81.68\%}  \\
  %Single-RNN  & \textbf{99.50\%}  & 99.51\%           & \textbf{81.83\%} & 81.58\%           \\
  %\end{tabular}
%    \begin{tabular}{l|c|c}
%               & \textbf{Level Selection}                        & \textbf{Reward Grounding}                      \\ \hline
%   IBM2 & 81.09\%           & 26.05\%           \\
%   Multi-NN    & 97.73\%           & 38.89\%           \\
%   Multi-RNN   & 99.33\%           & \textbf{82.02}\%           \\
%   Single-RNN  & \textbf{99.48\%}  & 80.54\%  \\
%   \end{tabular}
   {\small
   \begin{tabular}{lcc}
\toprule
              & Level Selection                        & Reward Grounding                      \\ \midrule
  IBM2 &     79.87\%      &     27.26\%      \\
  Multi-NN    &     93.51\%      &     36.05\%      \\
  Multi-RNN   & 95.71\%           & 80.11\%           \\
  Single-RNN  & \textbf{95.91}\%  & \textbf{80.46}\%  \\
\bottomrule
  \end{tabular}
  }
  \caption{Accuracy of 10-Fold Cross Validation (averaged over 3 runs) for each of the models on the AMT Dataset.}
  \vspace{-10pt}
  \label{table:results}
\end{figure}

Our evaluation tests the hypothesis that hierarchical structure improves the speed and accuracy of language grounding at multiple levels of abstraction.
We measure grounding accuracy and planning speed in simulation with a corpus-based evaluation,
%We perform a corpus-based evaluation in simulation and assess the speed and accuracy of our approach.
and demonstrate our system on a Turtlebot robot. 

\subsection{Mobile-Manipulation Robot Domain}
\label{sec:cw}
The Cleanup World domain~\citep{Junghanns1997SokobanAC,macglashan2015grounding}, illustrated in Fig.~\ref{fig:cw1}, is a mobile-manipulator robot domain that is  partitioned into rooms (denoted by unique colors) with open doors. Each room may contain some number of objects which can be moved (pushed) by the robot. 
%  The movement of the robot against a block in a particular direction will also move the block in that direction unless there is a wall or alternate object blocking the immediate path. 
This problem is modeled after a mobile robot that moves objects around, analogous to a robotic forklift operating in a warehouse or a pick-and-place robot in a home environment. We use an AMDP from \citet{gopalan17} for the Cleanup World domain, which imposes a three-level abstraction hierarchy for planning.
%A breakdown of the three-level abstraction can be found in the next paragraph.
 
The combinatorially large state space of Cleanup World simulates real-world complexity and is ideal for exploiting abstractions. At the lowest level of abstraction $L_0$, the (primitive) action set available to the robot agent consists of north, south, east, and west actions. Users directing the robot at this level of granularity must specify lengthy step-by-step instructions for the robot to execute. At the next level of abstraction $L_1$, the state space of Cleanup World %is drastically reduced to
only consists of rooms and doors. The robot's position is solely defined by the region (\textit{i.e.} room or door) it resides in. Abstracted actions are \emph{subroutines} for moving either the robot or a specific block to a room or door. It is impossible to transition between rooms without first transitioning through a door, and it is only possible to transition between adjacent regions;
%(\textit{i.e.} it would not be possible to directly move from the green room to the blue room in Fig.~\ref{fig:cw1} without first passing through the red room);
any language guiding the robot at $L_1$ must adhere to these dynamics. Finally, the highest level of abstraction, $L_2$, removes the concept of doors, leaving only rooms as regions; all $L_1$ transition dynamics still hold, including adjacency constraints.
%as deciding whether or not the robot may move to another room from its current room.
Subroutines exist for moving either the robot or a block between connected rooms. The full space of subroutines at all levels and their corresponding propositional functions are defined by \citep{gopalan17}. Fig.~\ref{table:Examples} shows a few collected sample commands at each level and the corresponding level-specific AMDP reward function.

\subsection{Procedure}
\label{sec:amt} 
%\vspace{-3mm}
We conducted an Amazon Mechanical Turk (AMT) user study to collect natural language samples at various levels of abstraction in Cleanup World.
%for the Cleanup World domain~\citep{Junghanns1997SokobanAC,macglashan2015grounding} (see Fig. \ref{fig:cw1}). 
Annotators were shown video demonstrations of ten tasks, always starting from the state shown in~Fig. \ref{fig:cw1}.
%using a single starting instance of the Cleanup World domain shown in Fig. \ref{fig:cw1}.
For each task, users provided a command that they would give to a robot, to perform the action they saw in the video, while constraining their language to adhere to one of three possible levels in a designated abstraction hierarchy: fine-grained, medium, and coarse. This data provided multiple parallel corpora for the machine translation problem of task grounding. We measured our system's performance by passing each command to the language grounding system and assessing whether it inferred both the correct level of abstraction and the reward function.  We also recorded the response time of the system, measuring from when the command was issued to the language model to when the (simulated) robot would have started moving. Accuracy values were computed using the mean of multiple trials of ten-fold cross validation. The space of possible tasks included moving a single step as well as navigating to a particular room, taking a particular object to a designated room, and all combinations thereof.

Unlike \citet{macglashan2015grounding}, the demonstrations shown were not only limited to simple robot navigation and object placement tasks, but also included composite tasks (\textit{e.g.} ``Go to the red room, take the red chair to the green room, go back to the red room, and return to the blue room'').
Commands reflecting a clear misunderstanding of the presented task, \textit{e.g.} ``please robot'', were removed from the dataset. Such removals were rare; we removed fewer than 30 commands for this reason, giving a total of 3047 commands.
%In order to not limit the potential variation in the provided language commands, only those commands which reflected a clear misunderstanding of the presented task were removed from the dataset. We removed fewer than 30 commands for this reason, giving us a total of 3047 commands; for example, ``please robot'' was removed as it is not clear what the robot should be doing.
Per level, there were 1309 $L_0$ commands, 872 $L_1$ commands, and 866 $L_2$ commands. The $L_0$ corpus included more commands since the tasks of moving the robot one unit in each of the four cardinal directions do not translate to higher levels of abstraction. 

%\stnote{Okay... I think this paragraph should be cut and moved to the robot response time section.  I put the previous comments in when I read it but then I was sort of expecting to see results... }
%\stnote{check this paragraph.}    \stnote{I think the previous few sentences could be unpacked more to explain why we are using these different planners.  I tried to give more context at the beginning of this paragraph which you should check.}
%to highlight the efficiency of using an abstraction hierarchy. %Specifically, we treat bounded RTDP \citep{McMahan2005BoundedRD} as the flat planner and test it separately as well as internal to the AMDP, measuring the time needed to solve the grounded tasks.
%Furthermore, we hypothesize that the addition of heuristics can further reduce the time needed for execution and test the hierarchical AMDP planner with and without the use of a Manhattan-distance heuristic.

\subsection{Robot Task Grounding}

 We present the baseline task grounding accuracies in Fig.~\ref{results:base} to demonstrate the importance of inferring the latent abstraction level in language. We simulate the effect of an oracle that partitions all of the collected AMT commands into separate corpora according to the specificity of each command. For this experiment, any $L_0$ commands that did not exist at all levels of the Cleanup World hierarchy were omitted, resulting in a condensed $L_0$ dataset of 869 commands. We trained multiple IBM2 and Single-RNN models using data from one distinct level and then evaluated using data from a separate level. Training a model at a particular level of abstraction includes grounding solely to the reward functions that exist at that same level. Reward functions at the evaluation level were mapped to the equivalent reward functions at the training level (\textit{e.g.} $L_1$ {\sf agentInRegion} to $L_0$ {\sf agentInRoom}). Entries along the diagonal represent the average task grounding accuracy for multiple, random 90-10 splits of the data at the given level. Otherwise, evaluation checked for the correct grounding of the command to a reward function at the training level equivalent to the true reward function at the alternate evaluation level.

Task grounding scores are uniformly quite poor for IBM2; however, IBM2 models trained using $L_0$ and $L_2$ data respectively result in models that substantially outperform those trained on alternate levels of data. It is also apparent that an IBM2 model trained on $L_1$ data fails to identify the features of the level. We speculate that this is caused, in part, by high variance among the language commands collected at $L_1$ as well as the large number of overlapping, repetitive tokens that are needed for generating a valid machine language instance at $L_1$. While these models are worse than what \citet{macglashan2015grounding} observed, we note that we do not utilize a task or behavior model. It follows that integrating one or both of these components would only help prune the task grounding space of highly improbable tasks and improve our performance. 

Conversely, Single-RNN shows the expected maximization along diagonal entries that comes from training and evaluating on data at the same level of abstraction. These show that a model limited to a single level of language abstraction is not flexible enough to deal with the full scope of possible commands. Additionally,  Single-RNN demonstrates more robust task grounding than statistical machine translation.  

%\stnote{Need to say that the NN models, as expected, are maximized along the diagonal, because they are training and testing on the same dataset.  Declare victory.  Also say why we are doing this (to compare against the full model).  I'm worried a reviewer might say it's not a fair comparison because we are testing on different training sets so of course it won't work, so we have to be clear why we are presenting it. }

%\danote{@SK: Did we agree that this was an issue with the delta and tau parameters?} \sknote{@DA - it's more a problem with the larger output space for L1 - especially for IBM 2, that tries to generate the reward function tokens, there's a lot of overlap between output reward functions that confuses things}

The task grounding and level inference scores for the models in Sec.~\ref{sec:lm} are shown in Fig.~\ref{table:results}. Attempting to embed the latent abstraction level within the machine language of IBM2 results in weak level inference. Furthermore, grounding accuracy falls even further due to sparse alignments and the sharing of tokens between tasks in machine language (\textit{e.g.} {\sf agentInRoom agent0 room1} at $L_0$ and {\sf agentInRegion agent0 room1} at $L_1$). The fastest of all the neural models, and the one with the fewest number of parameters overall, Multi-NN shows notable improvement in level inference over the IBM2; however, task grounding performance still suffers, as the bag-of-words representation fails to capture the sequential word dependencies critical to the intent of each command.  Multi-RNN again improves upon level prediction accuracy and leverages the high-dimensional representation learned by initial RNN layer to train reliable grounding models specific to each level of abstraction. 
%\stnote{Cut or give more detail; right now we don't know what the smaller exeprt-cleaned dataset is.  If we run it, need to motivate why we tried doing this and say more about how we cleaned the dataset and how many commands were in it.  I think we should probably cut.  We found that Multi-RNN outperformed all other models when presented with a smaller, expert-cleaned version of the collected AMT dataset; this improvement using a dataset with less variation across natural language and near perfect adherence to the underlying hierarchy suggests a successful use of the Multi-RNN heads to learn information specific to each abstraction level.} 
Finally, Single-RNN has near-perfect level prediction and demonstrates the successful learning of abstraction level as a latent feature within the neural model. By not using an oracle for level inference, there is a slight loss in performance compared to the results obtained in Fig.~\ref{results:baseRNN}; however, we still see improved grounding performance over Multi-RNN that can be attributed to the full sharing of parameters across all training samples allowing for positive information transfer between abstraction levels.

%stnote{Need to compare these results to table 5b and say that they are worse, because we are also inferring the level, which we provided to 5b (as a baseline).}

%\begin{itemize}
%	\item Go through table row-by-row, discuss each results
%    \item Talk about IBM 2 => Sparse alignments, too many shared tokens in reward function space (always agentInRoom, always rooms, etc.)
%    \item Talk about Multi-NN => Good results, but not great => unable to capture sequential word dependencies (go north, north, north, south is represented the same as go south, north, north, north)
%    \item Talk about Multi-RNN => Really good results, very solid level selection, etc. Draw attention to level specific models embedding prior information on hierarchy into data resulting in higher grounding accuracy than single-RNN
%    \item Talk about Single-RNN \sknote{SPECIFICALLY: Talk about why sharing parameter space and directly outputting joint conditional probability helps => POSITIVE INFORMATION TRANSFER between levels of abstraction, implicit level selection}
%    \item \sknote{If meaningful!} Go through T-SNE embeddings of natural language embeddings, show that level of abstraction words cluster together in space => namely, show distinction betwee (north, south, etc.), and (doors, rooms).
%    \danote{Alternatively, look at the LSTM cell activations to find distinct neurons firing for each abstraction level}
%\end{itemize}

\subsection{Robot Response Time}

Fast response times are important for
fluid human-robot interaction, so we assessed the time it would take a
robot to respond to natural language commands in our corpus.
%To assess response time,
We measured the time it takes for the system to
process a natural language command, map it to a reward function, and
then solve the resulting MDP to yield a policy so that the simulated
robot would start moving.
We used Single-RNN for inference since it was the most accurate grounding model,
and only correctly grounded instances were evaluated,
so our results are for $2634$ of $3047$ commands that Single-RNN got correct.

% We compared several different planners to
% solve the MDP.  First we use the state-of-the-art algorithm bounded
% RTDP \citep{McMahan2005BoundedRD} as the flat (non-hierarchical) MDP
% planning algorithm.  We compared it to the hierarchical AMDP
% planner of \citet{Gopalan2016PlanningWA}.  At the primitive level, the
% AMDP planner also requires a flat planner; again bounded RTDP was used
% to allow for comparable planning times.  Finally, we considered a
% version of the AMDP planner with a Manhattan-distance heuristic.
% While the flat planner technically allows for heuristics,
% distance-based heuristics are unsuitable for the composite tasks in
% our dataset.

We compared three different planners to solve the MDP:
\begin{itemize}
\item \textbf{BASE}: A state-of-the-art flat (non-hierarchical) planner, bounded real-time dynamic programming (BRTDP~\citep{McMahan2005BoundedRD}).
\item \textbf{AMDP}: A hierarchical planner for MDPs~\citep{gopalan17}. At the primitive level of the hierarchy ($L_0$), \textbf{AMDP} also requires a flat planner; we use \textbf{BASE} to allow for comparable planning times. Because the subtasks have no compositional structure, a Manhattan-distance heuristic can be used at $L_0$. While \textbf{BASE} technically allows for heuristics, distance-based heuristics are unsuitable for the composite tasks in our dataset. This illustrates another benefit of using hierarchies: to decompose composite tasks into subtasks that are amenable to better heuristics.
\item \textbf{NH} (No Heuristic): Identical to \textbf{AMDP}, but without the heuristic as a fair comparison against \textbf{BASE}.
\end{itemize}
We hypothesize \textbf{NH} is faster than \textbf{BASE} (due to use of hierarchy),
but not as fast as \textbf{AMDP} (due to lack of heuristics).
% \begin{itemize}
% \item \textbf{BASE}: A flat (non-hierarchical) planner for MDPs. We use bounded real-time dynamic programming (BRTDP) \lswnote{Cite; maybe mention before evaluation}.
% \item \textbf{AMDP}: A hierarchical planner for MDPs. Note that at the lowest level of the hierarchy (i.e., at the level of primitive actions), \textbf{BASE} is used for planning as well, with a Manhattan-distance heuristic \lswnote{Maybe this detail should be mentioned before the evaluation}.
% \item \textbf{NH} (No Heuristic): Same as \textbf{AMDP}, but without the heuristic. \textbf{NH} is designed to be a `fair' comparison against \textbf{BASE}, since neither use heuristics; the only difference is that \textbf{NH} has a task hierarchy. We hypothesize \textbf{NH} is faster than \textbf{BASE} (due to use of hierarchy), but not as fast as \textbf{AMDP} (due to lack of heuristics).
% \end{itemize}

\begin{figure}[t]
  \vspace{-10pt}
  \centering
  \begin{subfigure}{.48\linewidth}
  	\centering
    \includegraphics[width=1.1\linewidth]{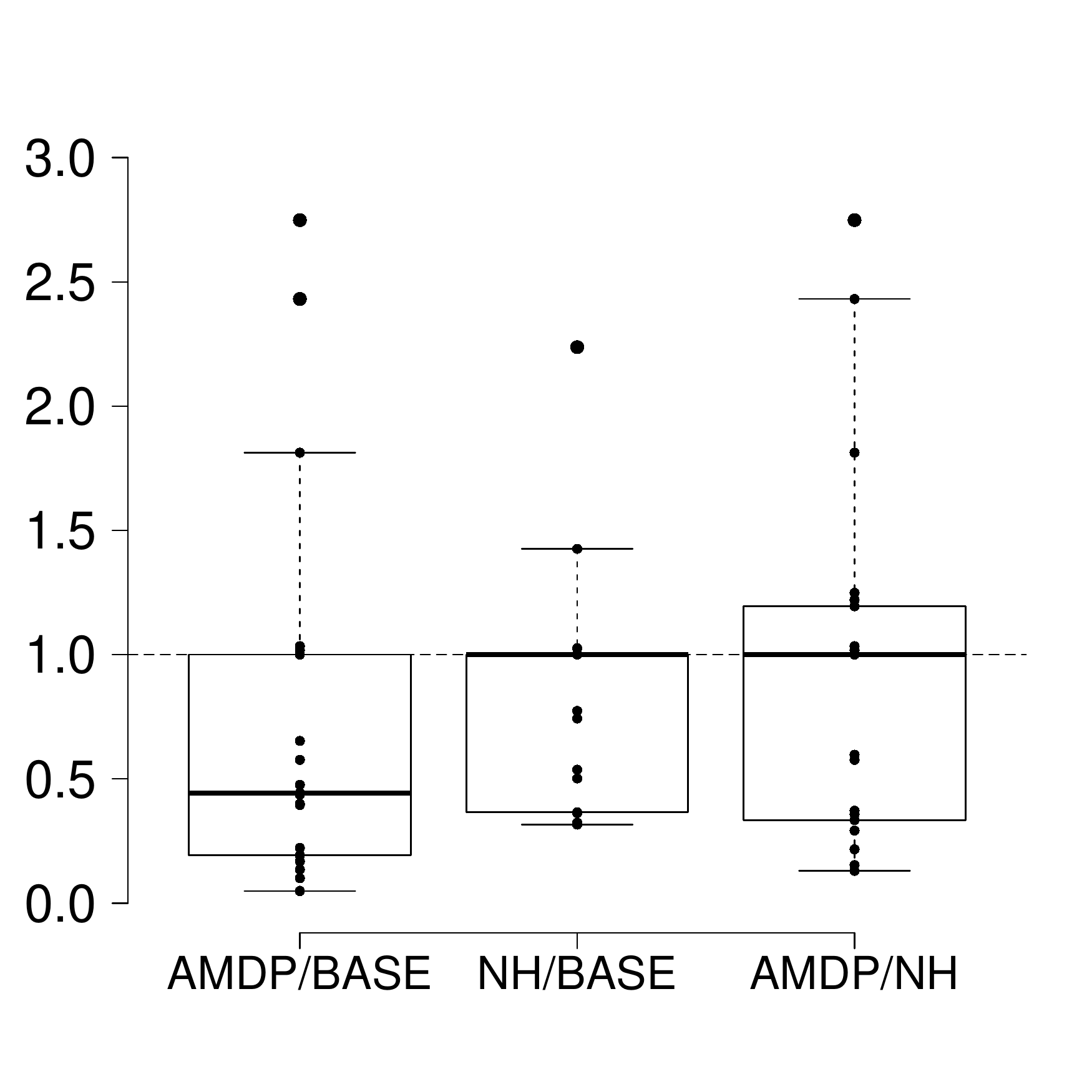}
    \vspace{-20pt}
    \caption{Regular domain ($2^{14}$ states)}
   	\label{fig:boxplot-reg}
  \end{subfigure}
  \hfill
  \begin{subfigure}{.48\linewidth}
  	\centering
    \includegraphics[width=1.1\linewidth]{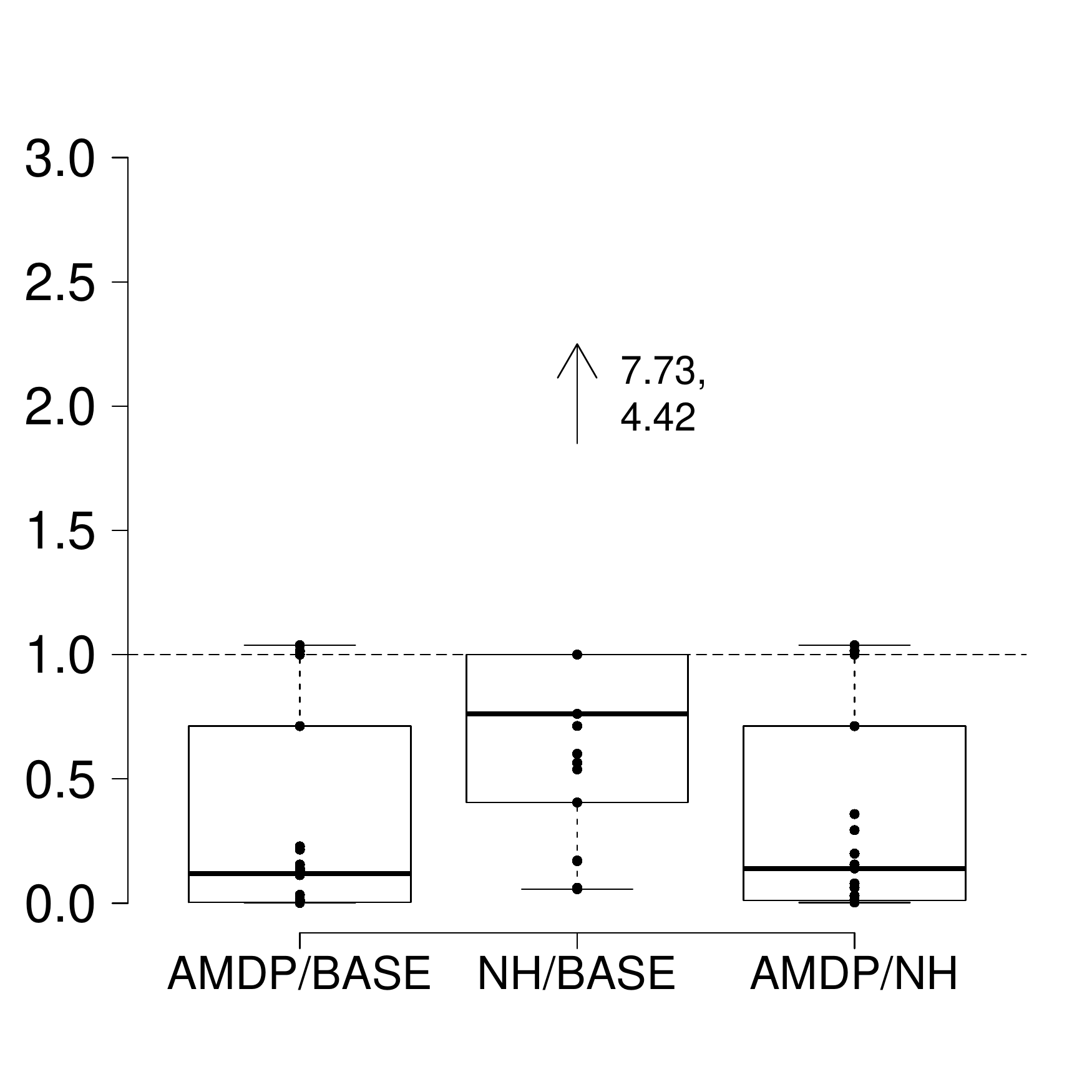}
    \vspace{-20pt}
    \caption{Large domain ($2^{18}$ states)}
   	\label{fig:boxplot-large}
  \end{subfigure}
  \hfill
  \caption{Relative inference $+$ planning times for different planning approaches on the same correctly grounded AMT commands. For each method pair, values less than $1$ indicate the method on the numerator (left of `$/$') is better. Each data point is an average of $1000$ planning trials.}
  \vspace{-10pt}
\label{fig:boxplot}
\end{figure}

Since the actual planning times depend heavily on the actual task being grounded
(ranging from $5 \text{ms}$ for {\sf goNorth} to $180 \text{s}$ for some high-level commands),
we instead evaluate the \emph{relative} times used between different planning approaches.
Fig.~\ref{fig:boxplot-reg} shows the results for all $3$ pairs of planners.
For example, the left-most column shows $\frac{\text{\textbf{AMDP} time}}{\text{\textbf{BASE} time}}$;
the fact that most results were less than $1$ indicates that \textbf{AMDP} usually outperforms \textbf{BASE}.
%\lswnote{Maybe give the actual fraction of which are better and/or equivalent.
%Consider also highlighting the fact that L0 groundings always use the same amount of time.}
Using Wilcoxon signed-rank tests, we find that each approach in the numerator is significantly
faster ($p < 10^{-40}$) than the one in the denominator, \textit{i.e.} \textbf{AMDP} is faster than \textbf{NH},
which is in turn faster than \textbf{BASE}; this is consistent with our hypothesis.
Comparing \textbf{AMDP} to \textbf{BASE}, we find that \textbf{AMDP} is twice as fast in over half the cases,
$4$ times as fast in a quarter of the cases, and can reach $20$ times speedup.
However, \textbf{AMDP} is also slower than \textbf{BASE} on $23\%$ of the cases;
of these, half are within $5\%$ of \textbf{BASE}, but the other half is up to $3$ times slower.
Inspecting these cases suggests that the slowdown is due to overhead from instantiating
multiple planning tasks in the hierarchy; this overhead is especially prominent in relatively small domains like Cleanup World.
Note that in the worst case this is less than a $2\text{s}$ absolute time difference.
%\stnote{This analysis might be a place to cut.  }
%\stnote{We note that the absolute magnitude of this slowdown in small problems is XX and will not effect the user experience much.}
%\stnote{Talk about slower cases too and why.  (overhead from the hierarchy.)}

From a computational standpoint, the primary advantage of hierarchy is space/time abstraction.
To illustrate the potential benefit of using hierarchical planners in larger domains,
we doubled the size of the original Cleanup domain and ran the same experiments.
Ideally, this should have no effect on $L_1$ and $L_2$ tasks,
since these tasks are agnostic to the discretization of the world.
The results are shown in Fig.~\ref{fig:boxplot-large},
which again are consistent with our hypothesis.
% \stnote{Sentence that states that these results are in line with our hypothesis.}
Somewhat surprisingly though, while \textbf{NH} still outperforms \textbf{BASE} ($p < 10^{-150}$),
it was much less efficient than \textbf{AMDP}, which shows that the hierarchy itself was
insufficient; the heuristic also plays an important role.
Additionally, \textbf{NH} suffered from two outliers,
where the planning problem became more complex because the solution was constrained
to conform to the hierarchy; this is a well-known tradeoff in hierarchical planning~\citep{Dietterich2000}.
%\lswnote{Emphasize that using hierarchy enables us to use better heuristics.}
The use of heuristics in \textbf{AMDP} mitigated this issue.
\textbf{AMDP} times almost stayed the same compared to the regular domain,
hence outperforming \textbf{BASE} and \textbf{NH} ($p < 10^{-200}$).
The larger domain size also reduced the effect of hierarchical planning overhead:
\textbf{AMDP} was only slower than \textbf{BASE} in $10\%$ of the cases,
all within $<4\%$ of the time it took for \textbf{BASE}.
Comparing \textbf{AMDP} to \textbf{BASE},
we find that \textbf{AMDP} is $8$ times as fast in over half the cases,
$100$ times as fast in a quarter of the cases,
and can reach up to $3$ orders of magnitude in speedup.
In absolute time, \textbf{AMDP} took $<1\text{s}$ on $90\%$ of the tasks;
in contrast, \textbf{BASE} takes $>20\text{s}$ on half the tasks.
%\stnote{What is the worst case?}

% \begin{figure*}[t]
%   \centering
%     \begin{tabular}{l}
%         \begin{tabular}[c]{@{}l@{}}
%         	Go south, south, south, south, west, south, west, west, west, north, north, north. \\
%         	Go down five spaces and then west until you hit the chair, push chair north five spaces.
%         \end{tabular} \\ 
%          \hline
%         \begin{tabular}[c]{@{}l@{}}
%         	Turn around and move forward. \\
%         	Move forward one step.  \\
%         \end{tabular} \\
%     \end{tabular}
%     \caption{Each row shows a pair of commands from the AMT dataset that are meant to ground to the same reward function whereas Single-RNN only finds the correct grounding for one command in the pair. The top command of the first row is grounded incorrectly due to no mention of the chair while the bottom command of the second row grounds incorrectly due to differing view on which direction to consider north.}
%     \label{table:confusion}
%   \hfill
% \end{figure*}

\subsection{Robot Demonstration}
Using the trained grounding model and the corresponding AMDP hierarchy, we tested with a Turtlebot on a small-scale version of the Cleanup World domain. To accommodate the continuous action space of the Turtlebot, the low-level, primitive actions at $L_0$ of the AMDP were swapped out for move forward, backward, and bidirectional rotation actions; all other levels of the AMDP remained unchanged.
%We modified the lowest level of the hierarchy, $L_0$, because the Turtlebot does not directly ground the actions of north, south, east, and west; instead the action set for $L_0$ is \emph{move forward}, \emph{move back}, \emph{turn clockwise} and \emph{turn counter clockwise}, which are longer sequences of commands inspired by the continuous commands possible on the Turtlebot.
%The \emph{move forward} and \emph{move back} commands move the robot forward or back by about \unit[10]{cm}. Further, the \emph{turn clockwise} and \emph{turn counter clockwise} change the robot's heading direction by $\pm90^{\circ}$. 
The low level commands used closed loop control policies, which were sent to the robot using the Robot Operating System \cite{quigley2009ros}.
%Further, the actions themselves were grounded on the robot using closed loop control.
%For the AMDP hierarchy all the other levels were kept the same, as they do not directly depend on the actions possible by the robot agent. 
%The planning within the AMDP hierarchy is done using BRTDP planners as mentioned previously.
Spoken commands were provided by an expert human user instructing the robot to navigate from one room to another. These verbal commands were converted from speech to text using Google's Speech API \citep{googleSpeechAPI} before being grounded with the trained Single-RNN model. The resulting grounding, with both the AMDP hierarchy level and reward function, fed directly into the AMDP planner resulting in almost-instantaneous planning and execution.
Numerous commands ranging from the low-level ``Go north'' all the way to the high-level ``Take the block to the green room'' were planned and executed using the AMDP with imperceivable delays after the conversion from speech to text. A video demonstration of the end-to-end system is available online: 
\href{https://youtu.be/9bU2oE5RtvU}{https://youtu.be/9bU2oE5RtvU}
%\footnote{\href{https://youtu.be/9bU2oE5RtvU}{https://youtu.be/9bU2oE5RtvU}}.
% Complex commands like "Take the block to the green room" were grounded by the planner at higher levels of abstraction, and simpler commands like "Go North" were grounded at the flat levels. We also have examples of low level commands like "Go down five steps and left four steps and up five steps" that get grounded to the reward functions at the lowest level of abstraction as they were specified in the language corresponding to lower levels. For all of this the planner functioned end to end with imperceivable delays after the speech command was received from the Speech API. The video is available online \footnote{\ngnote{fill up the video details!!!}} and is also submitted as part of the supplemental material.

\section{Discussion}
%\danote{The confusion matrix is quite large and I don't think there's much point in including it. Should the discussion include numbers based on that matrix or no? For example, XX\% of robot navigation commands at XX level failed due to XX\stnote{Failure analysis etc.  Specific examples that didn't work and why.} }
%Here we perform a brief error analysis of our best grounding model, Single-RNN, and examine a few sources of error using specific commands from the collected AMT dataset. 

%\stnote{This paragraph is more discussion... I would cut it and put it in the discussion section, after they have seen the results.}
%LSW% This paragraph seems redundant and more like a conclusion
% We present baseline results highlighting the difficulty of task grounding when modeling nuance in natural language, and go on to show dramatic improvement using our proposed methods. Furthermore, we show that while ignoring the underlying structure of natural language may result in better task grounding performance, the remaining planning problem cannot be solved efficiently for large domains. Together, our results demonstrate that we can maintain highly accurate task grounding as well as robust, efficient planning in complex environments. Finally, we demonstrate an end-to-end system using our approach deployed on a mobile robot.

Overall, our best grounding model, Single-RNN, performed very well, correctly grounding commands much of the time; however, it still experienced errors.  
At the lowest level of abstraction, the model experienced some confusion between robot navigation ({\sf agentInRoom}) and object manipulation ({\sf blockInRoom}) tasks. In the dataset, some users explicitly mention the desired object in object manipulation tasks while others did not; without explicit mention of the object, these commands were almost identical to those instructing the robot to navigate to a particular room. For example, one command that was correctly identified as instructing the robot to take the chair to the green room in Fig.~\ref{fig:cw1} is ``Go down...west until you hit the chair, push chair north...''
A misclassified command for the same task was ``Go south...west...north...'' These commands ask for the same directions with the same amount of repetition (omitted) but only one mentions the object of interest allowing for the correct grounding.   %\stnote{I see lots of different fonts being used for propositions: italics, fixed width (in the table), and regular font.  I vote for sans serif actually: {\sf agentInRoom}.  But anyway pick a font and be consistent.}
%\stnote{These sentences need example commands - it's not clear what is going wrong.  I wouldn't put them in the table but embed them in the text.  You don't have to include the whole command, use ``....'' to omit parts that aren't relevant.  Explain why the parts that are ommitted are important when at the wrong level.  Also can you explain more abstractly what is wrong?  I think it is saying ``The model experienced problems when commands mixed levels of abstractions, for example including actions at L0 and objects that are only defined at L1.''} 
Overall, 83.3\% of green room navigation tasks were grounded correctly while 16.7\% were mistaken for green room object manipulation tasks. 

Another source of error involved an interpretation issue in the video demonstrations presented to users. The robot agent shown to users as in Fig.~\ref{fig:cw1} faces south and this orientation was assumed by the majority of users; however, some users referred to this direction as north (in the perspective of the robot agent). This confusion led to some errors in the grounding of commands instructing the robot to move a single step in one of the four cardinal directions. Logically, these conflicts in language caused errors for each of the cardinal directions as 31.25\% of north commands were classified as south and 15\% of east commands were labeled as west.

Finally, there were various forms of human error throughout the collected data. In many cases, users committed typos that actually affected the grounding result (\textit{e.g.} asking the robot to take the chair back to the green room instead of the observed blue room). For some tasks, users often demonstrated some difficulty understanding the abstraction hierarchy described to them resulting in commands that partially belong to a different level of abstraction than what was requested. In order to avoid embedding a strong prior or limiting the natural variation of the data, no preprocessing was performed in an attempt to correct or remove these commands. A stronger data collection approach might involve adding a human validation step and asking separate users to verify that the supplied commands do translate back to the original video demonstrations under the given language constraints as in \citet{macmahon06}.

%\begin{itemize}
%\item Confusion between simple agentInRegion vs blockInRegion tasks; some Turkers explicity mention object while others simply provide the actions and acknowledge that the robot agent will push the object. This also caused issues between low and high level separation as objects were mentioned throughout.

%\item Occasional error in high level commands that caused correct proposition function but incorrect room choice. Variation in colors coupled with mentions of multiple rooms (e.g. must mention red room to cross into green room).
%\item Low level directional commands had some users confused due to directionality of robot agent icon. Agent faces south so some reversal of north-south. Similar issue for turning directionality in east-west instructions
%\item General errors in understanding of the hierarchy also exist in the raw data. Not sure if this is worth discussing; a critique of the abstraction hierarchy chosen suggesting importance of learning AMDPs?
%\end{itemize}
\section{Conclusion}

% \stnote{First paragraph:  summary of paper and main results.  In this paper we rpesented a system for interpreting natural language commands to a mobile-manipulator robot at multiple levels of abstraction.  Our system is the first model to use deep learning for language grounding on robots and outperforms previously publishe results.  We demonstrate that integrating with a hierarchical planner allows the system to interpret a much wider range of natural language commands which empowers human users.  Additionally we demonstrate with a robot evaluation that this system works in real-world environments.}
We presented a system for interpreting and grounding natural language commands to a mobile-manipulator robot at multiple levels of abstraction. To our knowledge, our system is not only the first work to ground language at multiple levels of abstraction, but also the first to use deep neural networks for language grounding on robots.
\lswnote{The rest of this section is new.}
Our proposed language-grounding models significantly outperform the previous state-of-the-art method for mapping natural language commands to reward functions. By explicitly considering the level of abstraction, our system can interpret a much wider range of natural language commands, as well as leverage an existing hierarchical planner for efficient planning and execution of robot tasks.
% We demonstrate that integrating such a language grounding system with a hierarchical planner allows for the specification and efficient execution of a wide range of robot tasks, fostering a very natural human to robot interaction.
%Finally, our Turtlebot evaluation demonstrates that this system works well in real-world environments and fosters natural human-robot interaction.
Finally, our Turtlebot evaluation demonstrates that this system works well in real-world environments and is an encouraging step towards seamless human-robot interaction.
\ngnote{I might weaken this claim, as we do not have any user studies to back this up}
\danote{@NG: Better? (Original commented out above)}
% We present baseline results highlighting the difficulty of task grounding when modeling nuance in natural language, and go on to show dramatic improvement using our proposed methods. Furthermore, we show that while ignoring the underlying structure of natural language may result in better task grounding performance, the remaining planning problem cannot be solved efficiently for large domains. Together, our results demonstrate that we can maintain highly accurate task grounding as well as robust, efficient planning in complex environments. Finally, we demonstrate an end-to-end system using our approach deployed on a mobile robot.

To achieve natural interaction with humans, robots must be able to interpret all possible natural language input. Therefore, we must weaken the constraints and assumptions we place on input from users.
To this end, we plan to extend our proposed models to handle natural language commands specified at a \emph{mixture} of abstraction levels. More generally, we should not allow an existing planning abstraction hierarchy to constrain our interpretation of language. In contrast, we can use the space of user inputs to \emph{inform} the learning of appropriate abstraction hierarchies, aiming to find structures that both match user language and are efficient to plan with.

We envision that our system is applicable to a large variety of real-world scenarios, particularly in environments where multiple levels of abstraction naturally occur, such as in surgical, manufacturing, and household robotics.

%Future work should extend this system via application to a large variety of real-world scenarios. Such a system would be effective in any environment where having multiple levels of abstraction make sense; for example, in surgical and household robotics.
%Additionally, it would be incredibly fruitful to extend the models proposed here to operate on natural language commands specified at a \emph{mixture} of abstraction levels to further reduce the constraints on natural language and facilitate more natural human-robot interaction. \lswnote{Weird phrasing in this final sentence. Consider cutting.} Alternate future work would relax the assumptions made in this work and allow for full variation in language or full variation in planning abstraction; one might, for example, learn abstraction hierarchies directly from the language. 

% In the future, we plan to explore more real-world scenarios, such as
% quad-rotor helicopters and an end-to-end user study. Additiona

% \begin{itemize}
% \item Mention future work: Treat a single natural language command as generated from multiple (or at least two different) levels of abstraction (e.g. Go the red room and take three steps to left).
% \end{itemize}

\section{Acknowledgements}
This work is supported by the National Science Foundation under grant number IIS-1637614, the US Army/DARPA under grant number W911NF-15-1-0503, and the National Aeronautics and Space Administration under grant number NNX16AR61G.

Lawson L.S. Wong was supported by a Croucher Foundation Fellowship.

%\lswnote{Change Gopalan et al. to be ICAPS paper.}

\bibliographystyle{plainnat}
\bibliography{references}

\end{document}

\maketitle

Here we provide details of the various language modeling approaches used for task grounding including a breakdown of the specific details of the IBM Model 2 Language Model, as well as each of the separate deep neural network models.

Recall that, given a natural language command $c$, we find the corresponding level of the abstraction hierarchy $l$, and the reward function $m$ that maximizes the joint probability of $l$, $m$ given $c$. Concretely, we seek the level of the state-action hierarchy $\hat l$ and the reward function $\hat m$ such that:
\begin{align} 
	\hat l, \hat m &= \arg \max_{l, m} Pr(l, m \mid c) \label{eq:CoreObj}
\end{align}

\subsection*{IBM Model 2 - Statistical Language Model:}

IBM2 is a generative model that solves the following objective, which is equivalent to Eqn.~\ref{eq:CoreObj} by Bayes' rule:
\begin{align}
%	\hat l, \hat m &= \arg \max_{l, m} Pr(l, m \mid c) \\
       \hat l, \hat m  &= \arg \max_{l, m} Pr(l, m) \cdot Pr (c \mid l, m) \label{eq:Noisy-Channel}
\end{align}

 In this equation, the first term, $Pr(l, m)$ can be treated as a distribution over the reward function space. We make the assumption that each $(l, m)$ tuple is distributed uniformly at random. Thus, the IBM2 learning objective simplifies to the following:
 \begin{align} \label{eq:IBMObjective}
 	\hat l, \hat m &= \arg \max_{l, m} Pr (c \mid l, m)
 \end{align}
This probability of a natural language command ($c$) given the reward ($m$) and level of abstraction ($l$) is then given by the following IBM2 equation:
\begin{equation} 
Pr(c | m, l) = \eta(n_c | n_m, l)\sum_a \prod_j^{n_c} \delta(a_j | j, n_c, n_m, l)\tau(c_j | m_{a_j}, l)
\end{equation}
where $\eta$, $\delta$, and $\tau$ are IBM2 specific parameters that are learned via the EM algorithm. $\eta(n_c \mid n_m, l)$ denotes the probability of generating a natural language command of length $n_c$ from a reward function of length $n_m$, and level $l$. The sum is defined over all possible alignments of natural language words to reward function tokens.  For computational efficiency, we approximate the sum by sampling from the set of possible alignments, following standard practice.

% $\delta(a_j|j, n_c, n_m, l)$ is the word alignment probability that the $j$th natural language word aligns with the $a_j$th reward function token for natural language command and reward function lengths $n_l$ and $n_m$ respectively; $\tau(l_j|m_{a_j}, l)$ is the word translation probability specifying the likelihood that natural language word $l_j$ is generated from the reward function $m_{a_j}$ at level $l$.

We take a standard approach to training our IBM2 using the EM algorithm with a ``bake-in'' period where the EM algorithm is run for a set number of iterations only for translation parameter ($\tau$) updates. We then learn follow with regular iterations of the EM algorithm where both the translation parameters ($\tau$) and the alignment parameters $(\delta)$ are updated. We estimate the length parameters ($\eta$) using Maximum-Likelihood estimation.

At inference time, to pick the $(l, m)$ tuple that maximizes the objective from Equation \ref{eq:IBMObjective} we calculate the IBM2 probability for every possible ($l, m$) combination, using the IBM2 as a reranker over the possible reward function translations. We find this gives significantly better results than beam-search decoding due to the relatively small size of the reward function space, as well the formulaic nature of each reward function string.

% Below, we step through the details of each of the different deep neural network architectures:

\subsection*{Multi-NN - Multiple Output Feed-Forward Network:}

A breakdown of the exact network transformations is as follows:
\begin{align*}
	\vec{e} &= \text{Lookup}(\mathbf{E}, \vec{c})  \\
    \vec{s} &= \text{ReLU}(\vec{e} \cdot \mathbf{W_s} + \mathbf{b}_s) \\
    \vec{t} &= \text{ReLU}(\vec{s} \cdot \mathbf{W_t^k} + \mathbf{b}_t^k) \\
    \vec{o} &= \text{Softmax}(\vec{t} \cdot \mathbf{W_t^k} + \mathbf{b}_t^k)
\end{align*}
Here, the layer specific weight and bias parameters are given by $\mathbf{W, b}$ respectively. Superscripts denote output-specific parameters and the ($\cdot$) operation denotes matrix-vector product. In order to produce high-dimensional, fixed-size representations of each word in the finite natural language vocabulary, the initial embedding layer contains a lookup matrix $\mathbf{E}$, trained via backpropagation with the rest of the model, where each row denotes a single word embedding. The embedding for all words in $\vec{c}$ are summed together according to their respective frequencies to produce an embedding for the full natural language command. All hidden layers employ the rectifier activation function (ReLU) whereas the final output layer produces a Softmax distribution over the output categories.

The fixed-size embedding is then passed through a neural network layer (shared across all outputs), with a ReLU non-linear activation, generating a hidden state vector $\vec{h}$. This hidden state vector is passed through an output-specific hidden layer, also with a ReLU activation, and finally an output-specific read-out layer, with a Softmax activation, to generate a probability distribution over the output categories. The loss is computed as the sum of the cross-entropy loss over the different outputs - namely, the computed loss for the level selection distribution, and each of the three different reward function distributions.

\subsection*{Gated Recurrent Units}

Both the Multi-RNN and Single-RNN models leverage Gated Recurrent Unit (GRU) cells, a specific type of Recurrent Neural Network cell. GRU Cells only maintain a single hidden state $h$, and the update rules are as follows:
\begin{align*}
	\vec{z}_t &= \sigma(\mathbf{W_z} \cdot \vec{x}_t + \mathbf{U_z} \cdot \vec{h}_{t - 1} + \mathbf{b}_z) \\
    \vec{r}_t &= \sigma(\mathbf{W_r} \cdot \vec{x}_t + \mathbf{U_r} \cdot \vec{h}_{t - 1} + \mathbf{b}_r) \\
    \vec{n}_t &= \text{Tanh}(\mathbf{W_h} \cdot \vec{x}_t + \mathbf{U_h} \cdot (\vec{r}_t \odot \vec{h}_{t - 1}) + \mathbf{b}_z) \\
    \vec{h}_t &= (\vec{1} - \vec{z}_t) \odot \vec{h}_{t - 1} + \vec{z}_t \odot \vec{n}_t
\end{align*}
Here, the ($\cdot$) operation denotes matrix-vector product, while the ($\odot$) operation denotes element-wise product. The intermediate vectors $\vec{z}, \vec{r}$ act as update and reset ``gates'' dictating how much of the hidden state should be overwritten with the new information in $x_t$. The parameters $\mathbf{W}, \mathbf{U}, \mathbf{b}$ are specific to the GRU cell, and are trainable via backpropagation along with the rest of the model. The hidden state $h$ is initialized as the zero vector at $t = 0$. 

\subsection*{Multi-RNN - Multiple Output Recurrent Network:}
We now give a detailed breakdown of the exact network transformations that make up the Multi-RNN:
\begin{align*}
	\vec{e}_1, \vec{e}_2 \ldots \vec{e}_n &= \text{Lookup}(\mathbf{E}, c_1, c_2 \ldots c_n) \\
    \vec{h} &= \text{GRU}(\vec{e}_1, \vec{e}_2, \ldots \vec{e}_n) \\
    \vec{s} &= \text{ReLU}(\vec{h} \cdot \mathbf{W_s} + \mathbf{b}_s) \\
    \vec{t} &= \text{ReLU}(\vec{s} \cdot \mathbf{W_t^k} + \mathbf{b}_t^k) \\
    \vec{o} &= \text{Softmax}(\vec{t} \cdot \mathbf{W_t^k} + \mathbf{b}_t^k)
\end{align*}
Again, layer parameters are given by $\mathbf{W, b}$, with superscripts denoting output-specific parameters. The loss is the same as that used by the Multi-NN model. 

\subsection*{Single-RNN - Single Output Recurrent Network:}

A detailed breakdown of the Single-RNN transformations are as follows:
\begin{align*}
	\vec{e}_1, \vec{e}_2 \ldots \vec{e}_n &= \text{Lookup}(\mathbf{E}, c_1, c_2 \ldots c_n) \\
    \vec{h} &= \text{GRU}(\vec{e}_1, \vec{e}_2, \ldots \vec{e}_n) \\
    \vec{s} &= \text{ReLU}(\vec{h} \cdot \mathbf{W_s} + \mathbf{b}_s) \\
    \vec{t} &= \text{ReLU}(\vec{s} \cdot \mathbf{W_t} + \mathbf{b}_t) \\
    \vec{o} &= \text{Softmax}(\vec{t} \cdot \mathbf{W_t} + \mathbf{b}_t)
\end{align*}
Note that these transformations are exactly the same as the Multi-RNN, with the sole exception that there is only a single output, rather than multiple. As there is only a single output, the new loss is just the cross-entropy loss of the predicted joint level-reward function distribution.